\documentclass[conference]{IEEEtran}
\IEEEoverridecommandlockouts

\usepackage{cite}
\usepackage{amsmath,amssymb,amsfonts}
\usepackage{algorithmic}
\usepackage{graphicx}
\usepackage{textcomp}
\usepackage{xcolor}
\usepackage{booktabs}

\usepackage{url}

\usepackage{xcolor}
\definecolor{orange}{RGB}{255,107,0}

\def\BibTeX{{\rm B\kern-.05em{\sc i\kern-.025em b}\kern-.08em
    T\kern-.1667em\lower.7ex\hbox{E}\kern-.125emX}}
\begin{document}

\title{Leveraging LLM Agents for Automated Optimization Modeling for SASP Problems: A Graph-RAG based Approach}

\author{
\IEEEauthorblockN{
    Tianpeng Pan\IEEEauthorrefmark{1}, 
    Wenqiang Pu\IEEEauthorrefmark{1},
    Licheng Zhao\IEEEauthorrefmark{1},
    Rui Zhou\IEEEauthorrefmark{1} 
}
\IEEEauthorblockA{
    \IEEEauthorrefmark{1}Shenzhen Research Institute of Big Data, The Chinese University of Hong Kong, Shenzhen, Guangdong, China\\
    Email: 224010189@link.cuhk.edu.cn, \{wpu,zhaolicheng,rui.zhou\}@sribd.cn
}
}

\maketitle

\begin{abstract}
Automated optimization modeling (AOM) has evoked considerable interest with the rapid evolution of large language models (LLMs). Existing approaches predominantly rely on prompt engineering, utilizing meticulously designed expert response chains or structured guidance. However, prompt-based techniques have failed to perform well in the sensor array signal processing (SASP) area due the lack of specific domain knowledge. To address this issue, we propose an automated modeling approach based on retrieval-augmented generation (RAG) technique,  which consists of two principal components: a multi-agent (MA) structure and a graph-based RAG (Graph-RAG) process. The MA structure is tailored for the architectural AOM process, with each agent being designed based on principles of human modeling procedure. The Graph-RAG process serves to match user query with specific SASP modeling knowledge, thereby enhancing the modeling result. Results on ten classical signal processing problems demonstrate that the proposed approach (termed as MAG-RAG) outperforms several AOM benchmarks. 
\end{abstract}


\section{Introduction}

Sensor Array Signal Processing (SASP) has experienced remarkable advancements over the past few decades~\cite{pesavento2023three}, which finds utility in a spectrum of applications, including telecommunications, radar, sonar, etc. Research within this field has encompassed areas such as beamforming, direction-of-arrival (DOA) estimation, primal user detection, source localization, etc. Over time, SASP has witnessed a paradigm shift from a predominantly parametric approach~\cite{krim1996two} to optimization methodologies~\cite{liu2023twenty,pesavento2023three}, leading to substantial advances in various application domains. Typically, SASP problems can be formulated as optimization problems, where mathematical formulations (objective functions and constraints) are established from the prior knowledge of the sensor system models and the final processing goal.

Traditionally, solving SASP problems necessitates the manual formulation and development of algorithms by human experts. However, the recent invention of Large Language Models (LLMs) demonstrates the potential to revolutionize SASP problem-solving. In particular, LLMs are capable of comprehending natural language inputs and generating logical sequences as responses, allowing users to describe the SASP problem and requirement in an intuitive way. Moreover, LLM has shown talent towards comprehension on mathematical equations~\cite{tang2024orlm,song2024towards,ahmaditeshnizi2023optimus}. This enables the automation of optimization model and algorithm suggestions, streamlining the process of finding effective solutions for diverse SASP problems. This approach is termed automated optimization modeling (AOM), which has great potential for immediate but reasonable solutions for a wide range of SASP applications.

Currently, AOM methods~\cite{ahmaditeshnizi2023optimus,tang2024orlm} predominantly utilize prompt engineering, including guiding LLMs to perform step-by-step reasoning~\cite{wei2022chain,yao2024tree,besta2024graph,kojima2022large,wang2022iteratively,gao2023pal,ahmaditeshnizi2023optimus} and deploying multi-agent systems to generate manually crafted response chains~\cite{xiao2023chain,chan2023chateval,talebirad2023multi,wu2023autogen}. This approach engages LLMs in constructing logical sequences for problem solving, emulating human cognitive processes~\cite{wei2022chain}. However, the inherent knowledge deficiencies present within LLMs has not yet been resolved. To clarify the knowledge referenced, the retrieval-augmented generation (RAG)~\cite{mialon2023augmented,lewis2020retrieval} method has recently been proposed. Notable performance improvements have been realized through the optimization of dataset structure~\cite{edge2024local} and the enhanced training of the retriever model~\cite{guu2020retrieval,ram2023context,weijia2023replug}.
However, despite these efforts, the SASP domain involves substantial domain-specific knowledge, limiting the success achieved by current AOM strategies.

To realize the potential of LLM-assisted AOM for SASP problem-solving, we introduce an automated modeling approach, which combines a multi-agent (MA) structure with a specific graph-based RAG (Graph-RAG) process. The MA structure is specifically tailored for the architectural complexities of AOM processes, following on principles of human expert's problem-solving logic. Each agent in the system is designed to tackle a segment of the problem, thereby decomposing a challenging mission into manageable sub-tasks~\cite{xiao2023chain}. The Graph-RAG component enhances this setup by matching user inputs with detailed domain modeling knowledge. This process mitigates the complexity of the AOM task, and further improves the performance by ensuring that only pertinent information is retrieved and utilized in modeling generation. Unlike traditional RAG, Graph-RAG organizes prior knowledge using a graph structure, making the retrieval process precise, crucial for fields like SASP where specific knowledge is needed \cite{lewis2020retrieval}. The proposed approach is termed as MAG-RAG. To evaluate it, we build a testing dataset, which includes 10 classical SASP problems along with recommended solutions. The experimental results indicate that MAG-RAG approach outperforms several AOM benchmarks. Meanwhile, several challenging issues are identified and discussed for further research.

\section{The Proposed AOM Approach}

\subsection{The MAG-RAG Pipeline}

\begin{figure}[!t]
    \centering
    \includegraphics[width=1\linewidth]{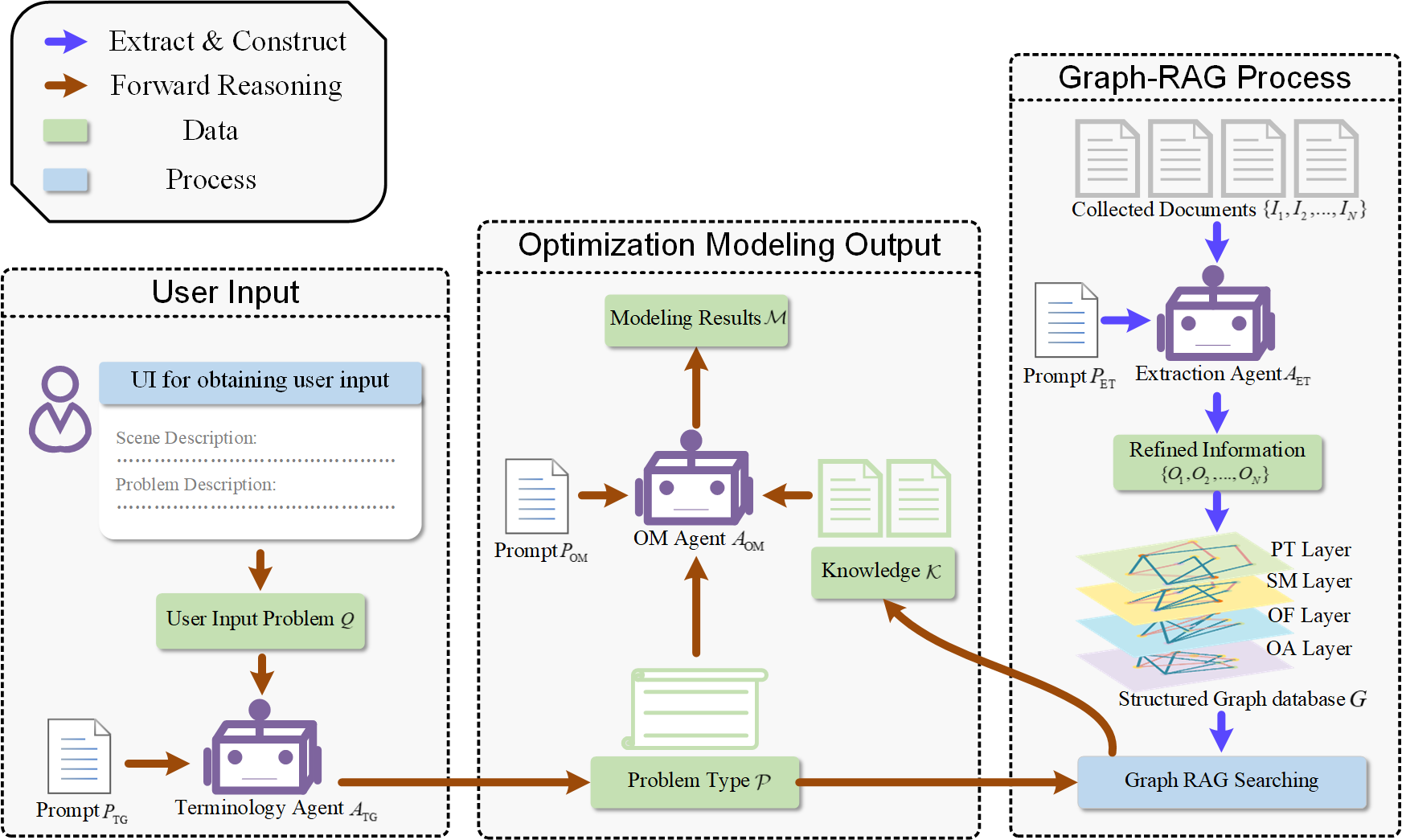}
    \caption{The overall workflow of the proposed method.}
    \label{fig:pipeline}
\end{figure}

The pipeline of the developed approach consists of two workflows as illustrated in Fig.~\ref{fig:pipeline}. The blue workflow illustrates the utilization of the Graph-RAG technique for constructing a knowledge database from domain-specific documents. This knowledge database can provide professional optimization modeling examples tailored to the user's query input (see example in Fig.~\ref{fig:intuition}). The other workflow in brown is the automated optimization modeling procedure, requiring the involvement of several agents. Before we formally introduce the pipeline structure\footnote{Agent in this work refers to one LLM, which takes specific prompt and other context as input. To make LLM agents being aware of their concrete responsibilities, each agent is provided a specially designed prompt (available at \url{https://github.com/advantages/MAG-RAG-for-SASP}) that outlines specific tasks, guidelines, and structured outputs.}, we specify three agents for AOM in the SASP domain, namely, Extraction Agent $A_{\textrm{ET}}$, Terminology Agent $A_{\textrm{TG}}$, and Optimization Modeling Agent $A_{\textrm{OM}}$. The role of each agent is explained as follows:

\textbf{Extraction Agent} $A_{\textrm{ET}}$. This agent receives raw documents $\left\{ I_{1},I_{2},\ldots,I_{N} \right\}$ in the SASP field as input, and strictly follows instructions from prompt $P_{\textrm{ET}}$ to extract knowledge $\left\{ O_{1},O_{2},\ldots,O_{N} \right\}$ critical for optimization modeling. Take $I_{i},~i \in \lbrack 1,\ldots,N\rbrack$ as an example, the key knowledge extraction process can be formulated as follows:

\begin{equation}
O_{i} = A_{\textrm{ET}}\left( P_{\textrm{ET}},I_{i} \right).
\end{equation}

\textbf{Terminology Agent} $A_{\textrm{TG}}$. This agent transforms original user input $\mathcal{Q}$ into terminological description $\mathcal{P}$. Given a specially designed prompt instruction $P_{\textrm{TG}}$, the extraction process is as follows:

\begin{equation}
\mathcal{P} = A_{\textrm{TG}}\left( P_{\textrm{TG}},\mathcal{Q} \right).
\end{equation}

\textbf{Optimization Modeling Agent} $A_{\textrm{OM}}$. This agent provides a complete modeling result $\mathcal{M}$ for $\mathcal{P}$ with reference to the extracted prior knowledge $\mathcal{K}$. Prior knowledge $\mathcal{K}$ is obtained from a Graph-RAG searching process (to be explained in Sec.~\ref{sec:ks}) with the query embedding of $\mathcal{P}$. The generation process of $A_{\textrm{OM}}$ can be formulated as follows:

\begin{equation}
\mathcal{M} = A_{\textrm{OM}}\left( P_{\textrm{OM}},\mathcal{K},\mathcal{P} \right).
\end{equation}

The pipeline of AOM is summarized as follows. Firstly, with the user-input query $\mathcal{Q}$ including the scene description, a Terminology Agent $A_{\textrm{TG}}$ converts unspecified user input query into a terminological problem description $\mathcal{P}$. Secondly, we retrieve top-k most relevant documents as reference knowledge $\mathcal{K}$ based on description $\mathcal{P}$. Finally, combining the terminology description $\mathcal{P}$ with reference knowledge $\mathcal{K}$, Agent $A_{\textrm{OM}}$ provides an answer $\mathcal{M}$ as the output. The retrieval technique in the second step is Graph-RAG which will be explained in the subsequent section.

\subsection{Graph-RAG Dataset Construction}
To assist LLMs with sophisticated SASP modeling, we construct a graph-based data base, where domain knowledge is extracted and represented as nodes, then weighted edges are formed among correlated nodes before knowledge searching. 

\textbf{Modeling Information Requirements:} Initially, $N$ domain knowledge documents $\left\{ I_{1},I_{2},\ldots,I_{N} \right\}$ are collected as raw property to construct the database. Though we anticipate the original documents to possess more information, they may also introduce information redundancy. Since not all the provided information contributes positively to the AOM task, excessive information imposes an additional burden on the modeling agent, which must first extract the essential information from the documents before proceeding with the subsequent modeling process. Besides, context limitation of LLMs~\cite{ding2024longrope,xiao2023chain} should also be seriously treated.

\begin{figure}[!t]
    \centering
    \includegraphics[width=1\linewidth]{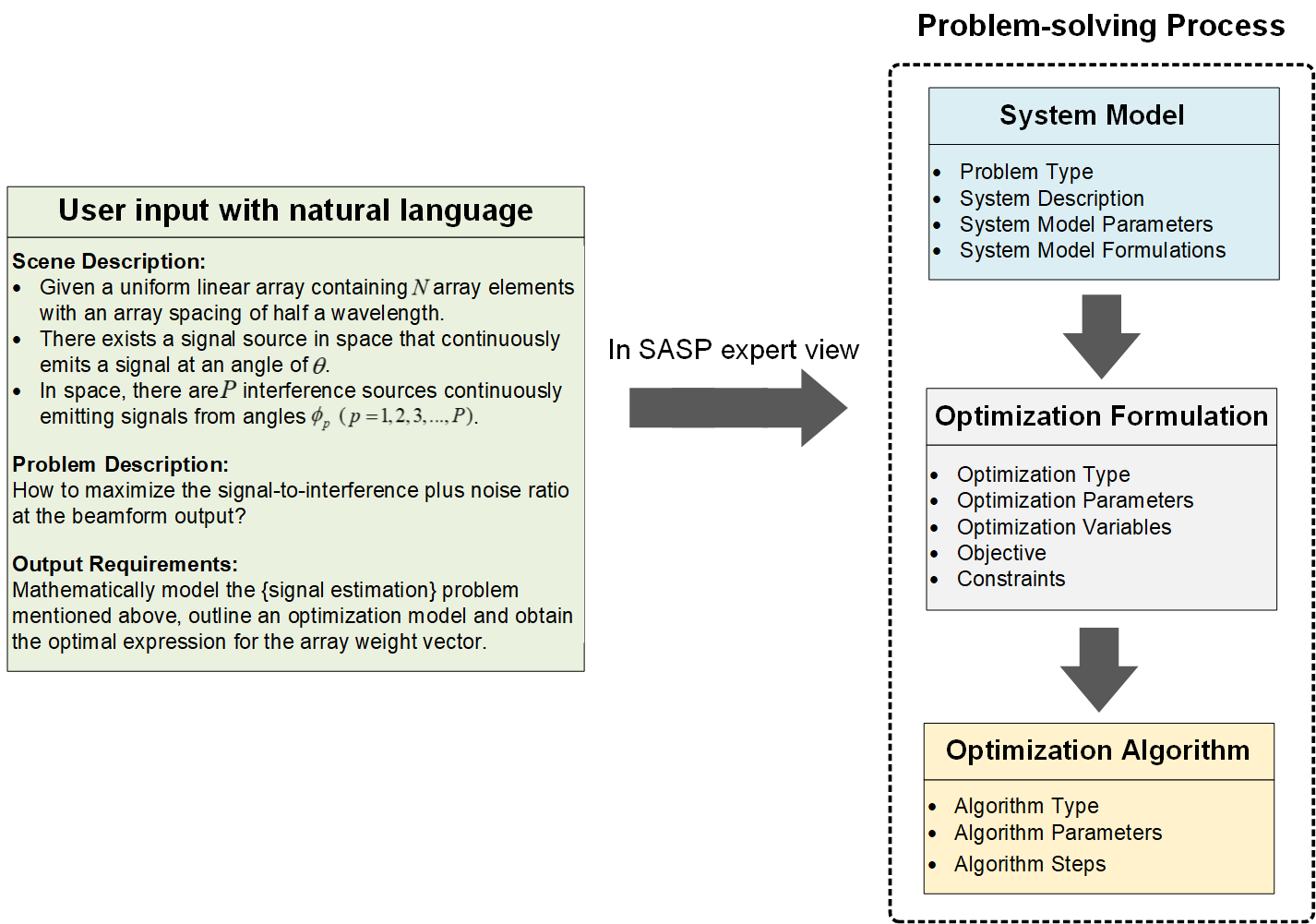}
    \caption{Human optimization modeling procedure for SASP problems.}
    \label{fig:intuition}
\end{figure}

\begin{figure}[!t]
    \centering
    \includegraphics[width=1\linewidth]{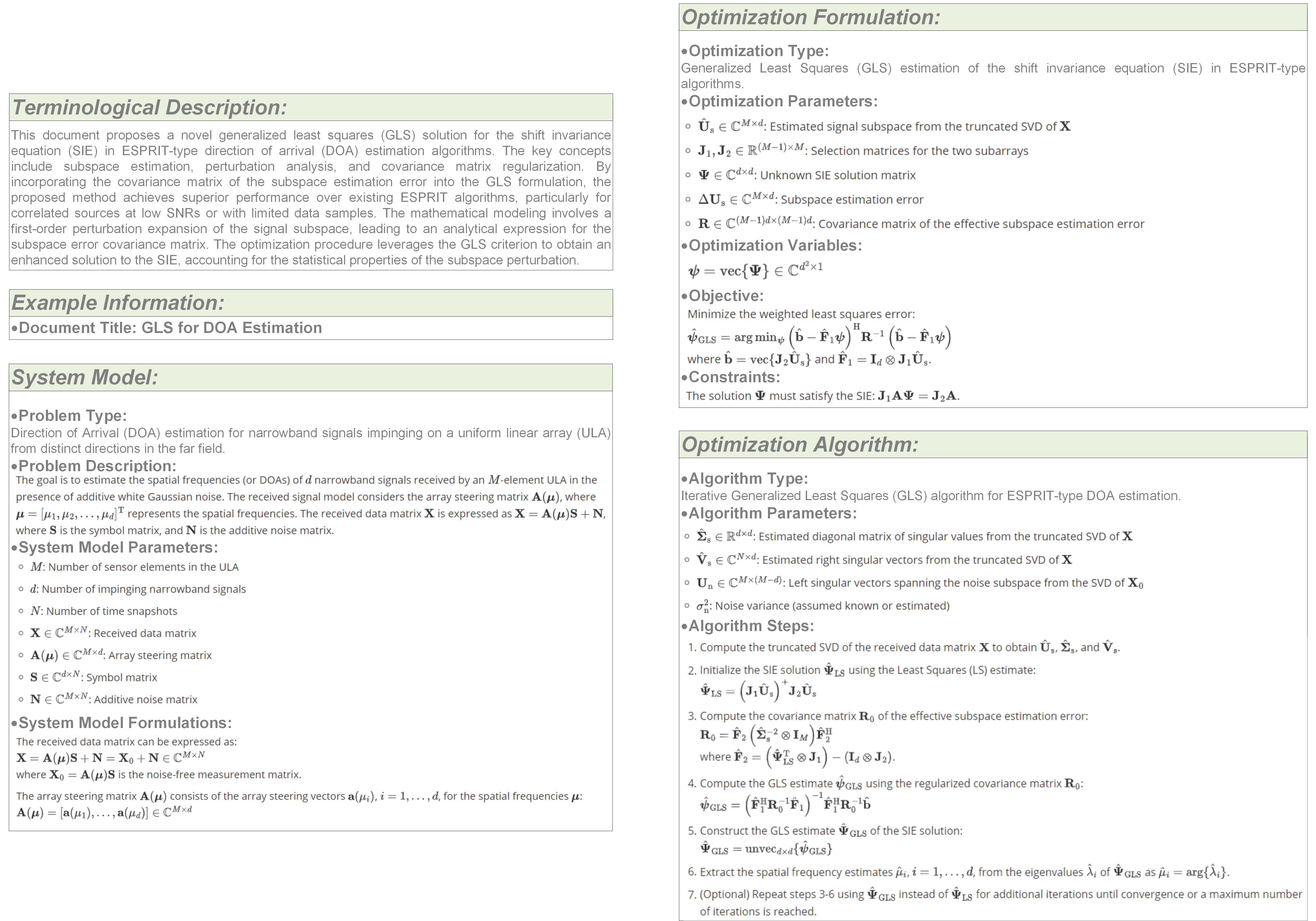}
    \caption{An example of the output generated by Example Extraction Agent}
    \label{fig:example}
\end{figure}

Thus, an Extraction Agent $A_{\textrm{ET}}$ is developed to distill the original documents into content pertinent to optimization modeling $\left\{ O_{1},O_{2},\ldots,O_{N} \right\}$. In order to conquer the uncontrollability of the LLM responses, we make it strictly follow the modeling mindset of human expert (as illustrated in Fig.\ref{fig:intuition}) with a target-definite prompt, where LLM response is instructed to consist of five parts: terminological description, example information, system model, optimization formulation and optimization algorithm. An example of the results is illustrated in Fig.~\ref{fig:example}.

\textbf{Dataset Construction of Graph-RAG:} The extracted content is formulated as a graph structure~\cite{velickovic2017graph,edge2024local}:

\begin{equation}
    G = \left( {V,E} \right),
\end{equation}
where $G$ denotes the graph dataset, $V$ and $E$ represent nodes and edges, respectively. Such a structure is chosen due to the fact that the graph structure is inherently endowed with a hierarchical process, allowing for a natural division of graphs into sub-graphs to capture the optimization modeling procedure. Additionally, graphs present superior flexibility for community clustering and efficient RAG searching. The intrinsic logic and hierarchical relationships inherent in graphs closely align with the optimization modeling procedure employed by human experts.

Concretely, a four-layer graph consisting of ``System Model (SM)" layer, ``Optimization Formulation (OF)" layer, ``Optimization Algorithm (OA)" layer and ``Problem Type (PT)" layer is constructed. For each $O_{i},~i \in \lbrack 1,\ldots,N\rbrack$, related content is traced and represented as nodes, which are subsequently allocated to the corresponding layers. Besides, nodes inherit the type of their respective layers, and are further assigned a ``keyword" attribute generated using all-MiniLM-L6-v2~\cite{he2024language} to serve as the searching query.

Two types of edges are developed according to the nodes' belonging entity: Nodes extracted from the same document are connected with ``single document (SD)" edges weighing 1.0, following the unoriented chain of ``PT-SM-OF-OA". Nodes extracted from different documents are linked with ``different documents (DD)" edges, with the edge weights representing their similarity. Specifically, for nodes $n_{i}$ and $n_{j}$, the embedding of each node's key words is generated as: $v_{i} = f_{e}\left(f_{k}\left( n_{i} \right) \right)$, $v_{j} = f_{e}\left(f_{k}\left( n_{j} \right) \right)$, where $f_{e}$ denotes the transformation from natural language to feature space using text-embedding-3-small~\cite{abdullahi2024retrieval}, and $f_{k}$ is the value extraction process from the node attribute ``keyword". Then, cosine similarity $s_{ij}$ is applied to calculate the relevance:

\begin{equation}
    s_{ij} = \frac{v_{i} \cdot v_{j}}{\left| \middle| v_{i} \middle| \middle| \times \middle| \middle| v_{i} \middle| \right|}.
\end{equation}
If $s_{ij}$ is greater than $\varepsilon$, a relationship between $n_{i}$ and $n_{j}$ is established, with $s_{ij}$ assigned as the value of ``similarity" attribute. 

\subsection{Knowledge Searching Using Graph-RAG}\label{sec:ks}

Knowledge searching process plays a vital role in Graph-RAG process. Due to the tendency of LLMs to utilize few-shot learning~\cite{cahyawijaya2024llms}, they prefer to draw on the given examples for responses. Furthermore, $A_{\textrm{TG}}$ produces content consistent with SASP terminologies, thus we employ ``PT'' layer for knowledge searching. For instance, considering a node $n_{p},p \in \lbrack 1,\ldots,N\rbrack$, the relevance of it to $\mathcal{P}$ can be determined as:

\begin{equation}
    s_{ep} = \frac{f_{e}\left( \mathcal{P} \right) \cdot f_{e}\left( f_{k}\left( n_{p} \right) \right)}{\left| \middle| f_{e}\left( \mathcal{P} \right) \middle| \middle| \times \middle| \middle| f_{e}\left( f_{k}\left( n_{p} \right) \right) \middle| \right|}.
\end{equation}
Then we manually produce a set $L$ that preserves the similarity of $n_p$ to $\mathcal{P}$:

\begin{equation}
    \left. L\leftarrow L \cup \left\{ {n_{p}:s}_{ep} \right\} \right..
\end{equation}

Finally, the nodes corresponding to the top-$k$ similarities in $L$ are selected. We build the knowledge $\mathcal{K}$ for $A_{\textrm{OM}}$ by concatenating the node content connected by ``SD" edges from these selected nodes. In this paper, we set $k=3$, by taking into account the context limitation and knowledge richness. 

\section{Experiments}

\subsection{Experimental Setup}
\textbf{Dataset: }We select ten classical SASP problems to evaluate the AOM performance, including transmitted beam pattern matching (Q1), cooperative sensing under ideal communication conditions (Q2), sensor placement (Q3), MIMO radar waveform design (Q4), direct positioning determination (Q5), DOA estimation (Q6), interference signal suppression (Q7), bearing-based localization (Q8), TOA-based localization (Q9) and TDOA-FDOA-based localization (Q10). For each issue, we finely select a number of documents containing standard modeling approaches to construct dataset SPAMR. The selected documents contain heuristic modeling strategies and optimization algorithms that can help researchers and LLMs improve the modeling process. The implementation code and evaluation dataset are available at \url{https://github.com/advantages/MAG-RAG-for-SASP}.

\textbf{Comparison Methods: }To fully evaluate MAG-RAG, we employ two external benchmarks. Pure MA refers to a pure agent logic chain for AOM, that the knowledge retrieval process of Graph-RAG is replaced by Knowledge Generation Agent $A_{\textrm{KG}}$. Pure LLM outputs the overall solutions for the input signal processing issue $\mathcal{Q}$ without any reference or prior knowledge.

\textbf{Metrics: }Five metrics granted different scores in overall 100\textquotesingle are adopted to evaluate the modeling results: Completeness (30\textquotesingle), Standardization (20\textquotesingle), Correctness (30\textquotesingle), Relevance (10\textquotesingle), Readability (10\textquotesingle).

\subsection{Performance Evaluation}

With the assistance of three human scientists specializing in SASP domain, we evaluate all the generated AOM results. The overall performances are shown in Table~\ref{tab:overall}.

\begin{table}[h]
\caption{Overall performance on different base LLMs. \\H: Haiku-3~\cite{anthropic2024claude}, S: Sonnet-3, G3.5: GPT-3.5, G4: GPT-4. \\D: pure LLM, G: MAG-RAG, T: pure MA}
\centering
\begin{tabular}{ccccccccccc}
\toprule
& Q1  & Q2  & Q3  & Q4  & Q5  & Q6  & Q7  & Q8  & Q9  & Q10 \\ 
\midrule
HD                  & 82  & 62  & 40  & 62  & 73  & 72  & 89  & 90  & 85  & 72  \\ 
HG                  & 70  & 60  & 70  & 75  & 64  & 81  & 61  & 63  & 42  & 46  \\ 
HT                  & 87  & 65  & 0   & 70  & 58  & 61  & 48  & 88  & \textbf{92} & 76  \\ 
SD                  & 75  & 21  & 42  & 71  & 68  & 75  & \textbf{91} & 91  & 90  & 84  \\ 
SG                  & 62  & 15  & \textbf{85} & \textbf{82} & \textbf{80} & \textbf{82} & 85  & \textbf{96} & \textbf{92} & \textbf{91} \\ 
ST                  & 70  & 40  & 35  & 79 & 71  & 80  & 80  & 92  & 87  & 81  \\ 
G3.5D               & 10  & 40  & 0   & 41  & 53  & 51  & 54  & 30  & 61  & 40  \\ 
G3.5G               & 75  & 40  & 44  & 49  & 64  & 73  & 46  & 38  & 43  & 46  \\ 
G3.5T               & 35  & 20  & 45  & 28  & 53  & 28  & 15  & 52  & 61  & 43  \\ 
G4D                 & 60  & 65  & 60  & 70  & 63  & 75  & 74  & 75  & 72  & 59  \\ 
G4G                 & \textbf{92} & 45  & 60  & 52  & 66  & 80  & 64  & 83  & 70  & 71  \\ 
G4T                 & 65  & \textbf{78} & 60  & 55  & 58  & 81  & 52  & 76  & 80  & 68  \\ 
\bottomrule
\end{tabular}
\label{tab:overall}
\end{table}

\begin{figure}[h]
    \centering
    \includegraphics[width=1\linewidth]{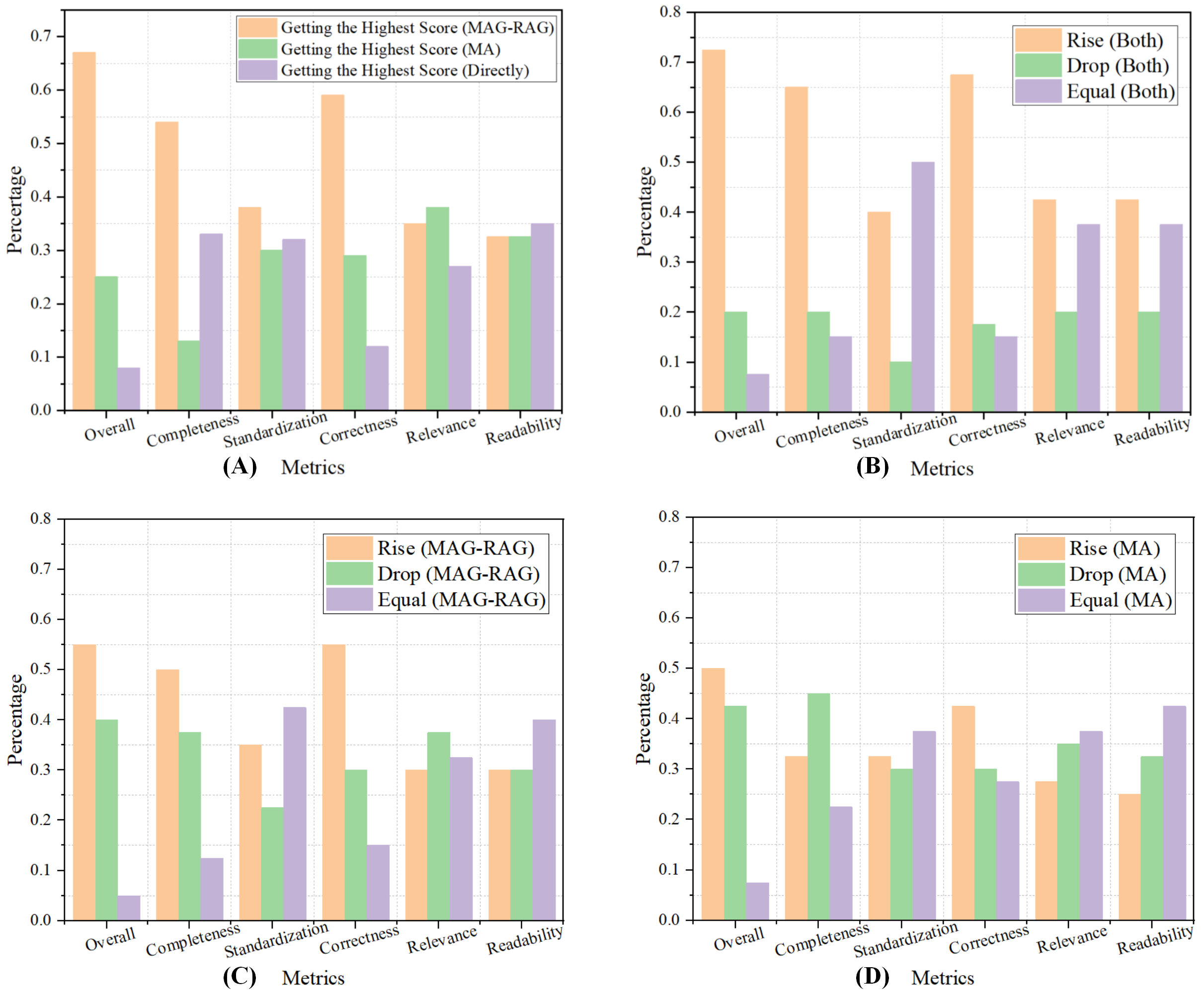}
    \caption{Statistical results on overall scores. (A) Frequency of different methods achieving the highest scores across various metrics. (B) Frequency of scoring gains (positive, negative and no gain) obtained with prior knowledge (both pure MA and MAG-RAG) compared to pure LLM. (C) Frequency of scoring gains obtained with MAG-RAG compared to pure LLM. (D) Frequency of scoring gains obtained with pure MA compared to pure LLM.}
    \label{fig:all_percentages}
\end{figure}

Three human scientists are responsible for evaluating Q1-Q3, Q4-Q6 and Q7-Q10, respectively. From Table.~\ref{tab:overall}, we discover that there remains a strong tendency in the given scores, where LLMs with MAG-RAG tend to achieve higher scores, despite the varied scoring preferences of the human scientists. Moreover, Sonnet-3 typically achieves better performance on modeling tasks across ten selected SASP problems. We statistically calculate the distributional properties of different metrics, and the results are shown in Fig.~\ref{fig:all_percentages}.

From Fig.~\ref{fig:all_percentages}(A), we observe that MAG-RAG achieves better results, occupying 67 percent on items getting the highest score, while pure MA and pure LLM approaches achieve only 25 percent and 8 percent, respectively. And in terms of completeness and correctness, MAG-RAG similarly far outperforms the comparison benchmarks, demonstrating that the knowledge extracted using a specially designed graph database has a positive effect on modeling. 

In viewing of standardization, relevance and readability, three methodologies perform similarly. This phenomenon is caused by the fact that LLM itself performs well in content formation, and the insertion of prior knowledge primarily aims to promote the optimization modeling.

From Fig.~\ref{fig:all_percentages} (B), we conclude that the utilization of prior knowledge for specific problems can indeed have a positive impact on modeling results, with the percentage of improved scores significantly exceeding that of declined scores. And in the cases where scores decreased, we statistically discover that four out of the eight samples had reduced scores originating from Q7. By referencing scientists' advice on Q7, directly invoking LLMs usually achieves higher results (89, 91, 54, 74), aligning with the expectations of different base LLMs. However, with the insertion of extracted prior knowledge, LLMs may additionally introduce incorrect constraints or miss key steps in algorithms.

In Fig.~\ref{fig:all_percentages} (C) and (D), we find that LLMs augmented with prior knowledge generally yield lower performance in terms of readability and contextual relevance compared to directly employing LLMs. Considering the auto-regression process in LLMs, this phenomenon may be attributed to the following factors: During the process of knowledge integration, the attention mechanism in Transformers often struggles to allocate attention weights appropriately, leading to biases in comprehension.

\section{Conclusion}
In this paper, we propose MAG-RAG approach for AOM, targeting SASP problems. We transform the AOM process into consecutive but separate parts, and based on this, a MA architecture is utilized to assign different sub-tasks into different LLMs. To enhance the efficiency, a graph-based RAG is adopted, where prior knowledge can be structurally stored and searched with more performance improvement.

Finally, we note that the proposed MAG-RAG mainly explores AOM for different SASP problems, while implicit relation between similar optimization algorithms is not sufficiently explored. Moreover, the inherent clustering strategy towards correlative SASP issues that may potentially enhance the knowledge searching efficiency is still unexplored.



\newpage

\bibliographystyle{IEEEtran}
\bibliography{ref}

\end{document}